\documentclass[11pt]{article}

\usepackage[]{acl}
\usepackage{booktabs}
\usepackage{multirow}
\usepackage{times}
\usepackage{latexsym}
\usepackage{booktabs}
\usepackage{subcaption}
\usepackage{float}
\usepackage[T1]{fontenc}

\usepackage[utf8]{inputenc}

\usepackage{microtype}

\usepackage{inconsolata}

\usepackage{graphicx}
\usepackage{tcolorbox}
\usepackage{listings}
\tcbuselibrary{breakable,skins}

\usepackage{tikz}
\definecolor{mygold}{HTML}{E3C800}
\definecolor{mygoldborder}{HTML}{B09500}

\DeclareRobustCommand*\goldcircled[1]{%
  \tikz[baseline=(char.base)]{
    \node[
      shape=circle,
      draw=mygoldborder ,   
      fill=mygold,   
      text=black,   
      thick,
      inner sep=0.0pt,
      minimum size=10pt, 
    ] (char) {#1};}}

\title{Automating Multi-Hop RAG Evaluation via TRIAD: From Context Extraction to Validated Dataset Generation\thanks{Accepted at the 19th International Natural Language Generation Conference (INLG 2026).}}

\author{Lorenz Brehme \\
  University of Innsbruck \\
  Department of Computer Science \\
  Innsbruck, Austria \\
  \texttt{lorenz.brehme@uibk.ac.at} \\\And
  Adam Jatowt \\
  University of Innsbruck \\
  Department of Computer Science \\
  Innsbruck, Austria \\
  \texttt{adam.jatowt@uibk.ac.at} \\}

\begin{document}
\maketitle
\begin{abstract}
Recent advances in LLMs and the adoption of RAG systems in industry have created a need for domain-specific question-answer datasets that can assess RAG performance on proprietary data. Existing datasets, such as HotpotQA, challenge current RAG systems on Wikipedia-based knowledge, but they cannot be transferred directly to domain-specific settings. A comprehensive evaluation of RAG system quality requires both multi-hop queries and unanswerable questions.
This paper introduces TRIAD, a three-stage automated dataset generation approach. First, it generates question--answer (QA) pairs for the domain-specific knowledge base of a RAG system. Second, a validator checks each QA-pair in a feedback loop. Third, the QA pairs are extended with relevance-labeled context documents for downstream evaluation.
We evaluate this approach against the established MuSiQue and HotpotQA datasets. The results show that the generated dataset exhibits similar performance trends across different RAG setups, while human validation indicates that the questions are suitable for evaluating a domain-specific RAG system. The code used to generate the dataset and all validation results are available in our GitHub repository (\url{https://github.com/lorenzbrehme/triad}).
\end{abstract}

\section{Introduction}

Retrieval-Augmented Generation (RAG) has evolved rapidly \cite{brehme_retrieval-augmented_2026}, leading to numerous approaches to improve and evaluate RAG systems \cite{yu_evaluation_2024}. Evaluation frameworks such as RAGAS \cite{es_ragas_2023} assess both the retriever and generator components.
Retriever evaluation measures whether the system returns relevant context documents for a query by comparing retrieved contexts with predefined relevance labels or by using relevance judgments from human annotators or LLMs.
Generator evaluation focuses on the quality of the RAG answers. Short answers can often be evaluated with exact-match metrics, while longer answers require assessment by humans or LLM judges \cite{ding_vera_2024, es_ragas_2023, saad-falcon_ares_2024}. Important criteria include correctness \cite{yang_crag_2024} and faithfulness to the retrieved context \cite{liu_cofe-rag_2024, ru_ragchecker_2024}.
These evaluations rely on QA datasets containing at least questions, and often additional attributes such as reference answers or labeled relevant contexts \cite{yang2018hotpotqa}. A key distinction is between single-hop questions, answerable from one document, and multi-hop questions, which require combining information from multiple sources \cite{yang2018hotpotqa}. Multi-hop evaluation matters because RAG systems must combine information from multiple sources to answer questions effectively \cite{tang_multihop-rag_2024, brehme_evaluating_2026}.

There are several existing datasets for multi-hop questions, including datasets with context relevance labels, such as HotpotQA \cite{yang2018hotpotqa} and MuSiQue \cite{trivedi2021musique}. One limitation of these datasets is that they are restricted to specific domains. In practical RAG evaluation, however, the underlying knowledge base is often proprietary and not publicly available. As a result, public benchmarks cannot fully test system behavior on the target corpus, which necessitates the creation of a custom QA set.
One option is to create this dataset manually \cite{schimanski_climretrieve_2024}. This approach is labor-intensive and costly because it requires expert annotators. Another option is to generate the dataset with an LLM. Many frameworks already address the generation of synthetic datasets \cite{pu_customized_2024}. However, most of them focus on relatively simple single-hop queries, which do not fully reflect RAG requirements \cite{krishna_fact_2024}. Other frameworks support multi-hop question generation, but they often require extensive preprocessing, for example by constructing a knowledge graph from documents \cite{lee_grade_2025, chen_semantic_2025} or by labeling each document in advance \cite{park_dochop-qa_2025, filice_generating_2025}. 
In addition, many existing approaches do not include unanswerable questions. These are important for evaluating whether a RAG system hallucinates an answer or correctly declines to answer when the available context is insufficient \cite{ravi_lynx_2024}.

The difficulty of creating QA datasets for evaluating RAG systems is also reflected in current industrial adoption: although companies are increasingly deploying RAG technology, they still struggle with reliable system evaluation \cite{brehme_retrieval-augmented_2026}. In this work, we propose TRIAD, a three-stage approach for automatically generating a multi-hop QA dataset with labeled context documents and unanswerable questions. Our approach leverages the existing vector database used by the RAG system, requires no complex preprocessing, and directly generates a QA dataset that can be used for evaluation, enabling a directly deployable evaluation setup. This enables the creation of evaluation datasets for domain-specific RAG corpora, rather than limiting evaluation to predefined benchmark knowledge. We validate the approach by manually assessing both the quality of the generated questions and the behavior of the validator, and by comparing the resulting datasets with the established HotpotQA and MuSiQue benchmarks to determine whether they exhibit similar performance trends across seven different RAG setups. This study serves as a foundational validation of the proposed pipeline, demonstrating its ability to generate reliable evaluation datasets while preserving meaningful performance distinctions between different RAG systems. The resulting datasets demonstrate high quality, exceeding 90\% in both answerability and correctness. Compared with the existing benchmarks, the RAG systems achieved better performance on the generated questions, but the datasets still exhibited similar performance trends across the tested RAG setups. In addition, the generated questions are not restricted to a specific knowledge source and are easily adaptable.
All code and data from the experiments are available in our GitHub repository \cite{githubGitHubLorenzbrehmetriad}.

\section{Related Work}

Multi-hop questions require combining different pieces of knowledge to produce an answer \cite{mavi_multi-hop_2024}. Several datasets have been introduced for this purpose, including MuSiQue \cite{trivedi2021musique}, 2WikiMultiHopQA \cite{ho_constructing_2020}, and HotpotQA \cite{yang2018hotpotqa}. Different question types were introduced by \citet{yang2018hotpotqa}, including comparison questions, which compare two entities, and bridging questions, in which one bridge entity connects information across two documents.

Another important concept in multi-hop question answering is context relevance. This refers to whether a context is relevant for answering a given question \cite{saad-falcon_ares_2024}. In datasets such as HotpotQA and MuSiQue, each question is associated with a set of context documents labeled as relevant or non-relevant. These labels can be used for retriever evaluation \cite{brehme2025SLR}. However, these datasets are manually created and tailored to their specific source corpora. As a result, they do not evaluate RAG-specific knowledge and are therefore not directly suitable for evaluating RAG systems operating on domain-specific corpora.
To address this limitation, QA datasets can be generated automatically with an LLM.

Several frameworks have already addressed automatic generation of such questions \cite{park_dochop-qa_2025, tang_multihop-rag_2024}. Many of these approaches include a preprocessing step in which a knowledge graph is constructed \cite{lee_grade_2025, chen_semantic_2025}. This graph is then used to identify related entities and supporting documents for question generation. Another approach iteratively rewrites a question by adding new documents to create a multi-hop question \cite{hwang_explainable_2024}.
A related line of work aims to trigger RAG systems with multiple possible answers in order to create ambiguous questions \cite{ji_deepambigqa_2025}. This makes it possible to evaluate RAG systems in settings involving incomplete or uncertain information.
Another aspect of evaluating a RAG system is how it reacts to questions that are unanswerable. This issue was addressed by \citet{liu_investigating_2026}, who first generated multi-hop queries and then modified them with public knowledge obtained through web searches. However, this approach is limited to general knowledge and cannot be transferred to domain-specific settings, where relevant information is unavailable from external sources.

To obtain a multi-hop dataset without such complex preprocessing and to include irrelevant contexts for unanswerable questions, we introduce our approach for generating multi-hop QA sets for complex questions. In addition, we define different question types to trigger different retriever behaviors.

\section{TRIAD Methodology}

\begin{figure*}[t]
  \centering
  \includegraphics[width=\textwidth]{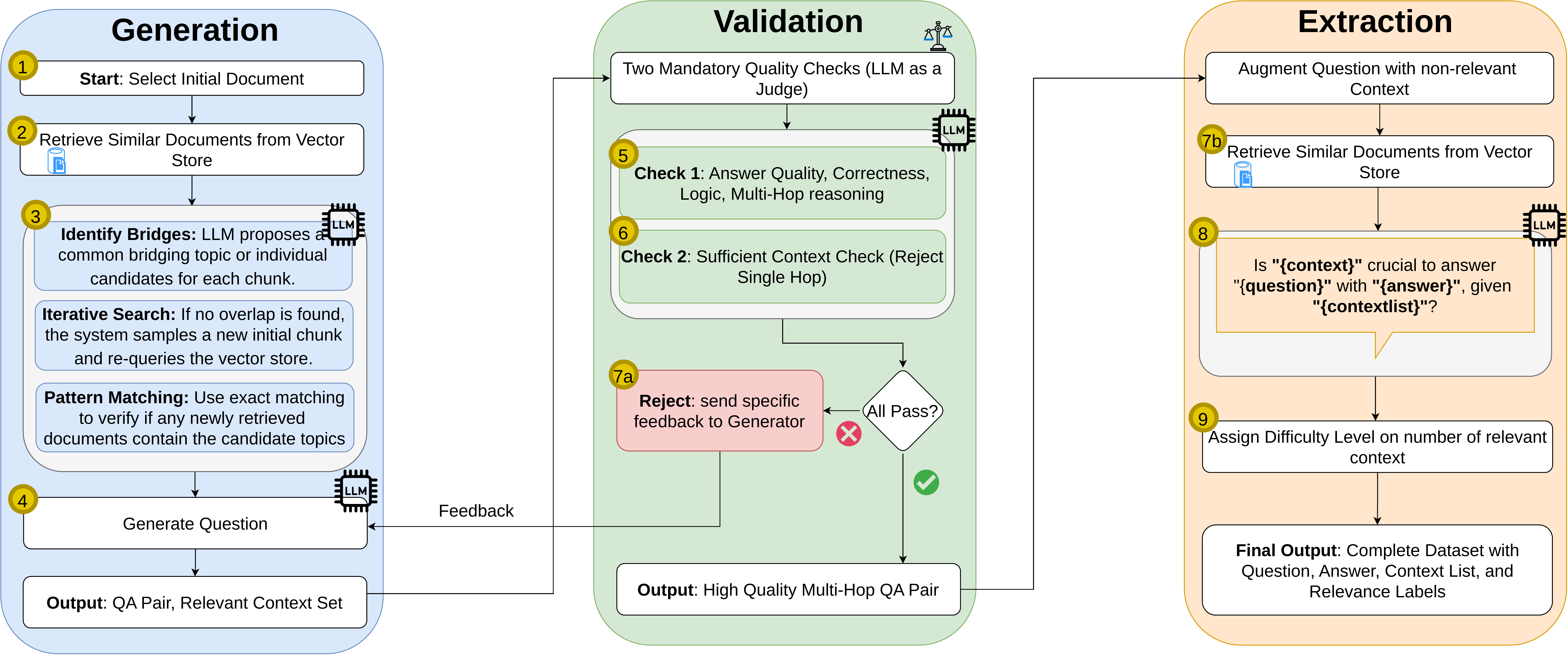}
  \caption{Overview of the TRIAD approach}
\label{fig:triad_appraoch}
\end{figure*}

In this section, we introduce TRIAD, a three-stage approach for generating multi-hop questions with labeled context documents (cf. Figure \ref{fig:triad_appraoch}). It consists of generation, validation, and extraction: questions are first created from source content and type requirements, then validated for quality and multi-hop reasoning, and finally paired with relevant and non-relevant documents to form a labeled context list.

\subsection{Generation}
The generation stage creates questions based on documents required to answer them.
First, an \textit{initial document is selected} from the RAG corpus \goldcircled{1}. This can be done manually, or the document can be sampled at random.
The initial document is then used to \textit{retrieve a set of additional documents} by querying the vector store for similar entries \goldcircled{2}. Once this set is available, a \textit{bridging topic must be identified} to connect the chunks and support QA generation. For this step \goldcircled{3}, the LLM is prompted to produce a bridging topic for the retrieved chunks. If no bridging topic exists, it returns an empty string. In addition, the LLM outputs a list of potential bridging topics for each chunk.
This list is used when no common bridging topic is found across all chunks. In that case, another initial chunk is selected, and the top-$n$ similar documents are retrieved again. We then check, via exact matching, whether each retrieved document contains the candidate bridging topic. We retain only configurations that satisfy the required number of chunks sharing the same bridging topic. If no valid configuration is found, the next chunk from the initial candidate list is used and the process is repeated. If no common topic can be identified, the pipeline restarts. Otherwise, we obtain a document set and a corresponding bridging topic.

The next step is \textit{question generation} \goldcircled{4}. For this step, an LLM is prompted with the bridging topic, the document set, the question type, and the target role the question should fulfill. The LLM outputs both the question and the corresponding answer. \textit{This results in a sample that contains the question, the answer, the set of relevant context documents, and the associated type and role labels}.

\subsubsection{Question Types}
To obtain a diverse set of examples, we define three multi-hop question types—\textit{comparison}, \textit{intersection}, and \textit{composition}—and two additional cross-cutting question categories, \textit{binary} and \textit{temporal} (cf. Table \ref{tab:multi_hop_types}). We additionally consider \textit{unanswerable questions} as a separate category.
\paragraph{Multi-hop question types.}
The three multi-hop question types are characterized by two distinct retrieval patterns: parallel retrieval and sequential retrieval.

The first pattern is \textbf{parallel retrieval}. In this setting, the question already contains enough information to retrieve all relevant chunks directly, because all key entities are explicitly mentioned.
For example: ``Which city has a larger population, London or New York?'' In this case, the retriever can directly use ``London'' and ``New York'' as query anchors and retrieve the relevant chunks in parallel. This parallel retrieval pattern is used by both the \textit{comparison} and the \textit{intersection} question type. The comparison type contrasts multiple entities to derive an answer, while the intersection type filters or aggregates entities, e.g., ``Which actor appeared in both *Inception* and *Interstellar*?''

The second retrieval pattern is \textbf{sequential retrieval}. Here, the query is structured such that an intermediate entity must first be identified from part of the question. This entity has to be retrieved before the remaining part of the question can be resolved, and only then can the next chunk be retrieved to produce the final answer. This pattern includes the \textit{composition} question type. Consider the question: ``What is the capital of the country of birth of Person~A?'' First, the RAG system must retrieve information about Person~A to identify the birthplace. It then uses this birthplace to retrieve information about that location and determine its capital city. This reasoning process cannot be completed by retrieving all chunks in a single step. It requires sequential retrieval, which is not supported by a simple single-step RAG architecture.

\paragraph{Cross-cutting question categories.}
In addition to these retrieval patterns, we define two cross-cutting question categories. One category is \textit{binary questions} (i.e., yes/no questions), which can be applied to all retrieval patterns. The other category is \textit{temporal questions}, which require temporal reasoning within the question.
Temporal questions are classified in a later step. Specifically, we use an LLM to determine whether each question requires temporal reasoning.

\paragraph{Unanswerable questions.}
Finally, we include \textit{unanswerable questions} as an additional question category. In these cases, the question cannot be answered using the documents included in the RAG corpus. The expected system behavior is to abstain (e.g., by stating that the question cannot be answered) rather than hallucinating an incorrect answer.
This question type does not include retrieval evaluation. Instead, we simulate the retriever by directly providing contexts, which are later extended by the extraction stage. We then provide the generator only with non-relevant contexts, or with only one partially relevant context, to ensure that the question remains unanswerable based on the provided documents. This design reduces the risk that another context in the corpus contains the answer but is missed during generation. Verifying unanswerability with respect to the entire corpus would require substantial additional effort.

\begin{table*}[t]
\centering
\small
\begin{tabular}{p{1.4cm}p{1.4cm}p{5.7cm}p{5.3cm}}
\toprule
\textbf{Question Type} & \textbf{Retrieval Pattern} & \textbf{Description} & \textbf{Example} \\
\midrule
Comparative & Parallel & Compare two or more entities that can be retrieved independently & Which city is larger, Paris or Berlin? \\
\midrule
Intersection & Parallel & Retrieve multiple entities independently, then filter or aggregate them to satisfy constraints & Who acted in both Film~A and Film~B? \\
\midrule
Composition & Sequential & Retrieve information in sequence, where the first retrieval leads to the next & What is the capital of the birthplace of Person~C? \\
\midrule
Binary & Sequential / Parallel & Questions with a yes/no answer & Is the capital of the birthplace of Person~C Town~A? \\
\midrule
Temporal & Sequential / Parallel & Require temporal reasoning across one or more facts & Who was the U.S. president when the company that developed Android was founded? \\
\midrule
Unanswerable & \multicolumn{1}{c}{\textemdash} & Questions that cannot be answered from the provided corpus; the model should abstain & \multicolumn{1}{c}{\textemdash} \\
\bottomrule
\end{tabular}
\caption{Classification of multi-hop question types by retrieval pattern.}
\label{tab:multi_hop_types}
\end{table*}

\subsection{Validation}
The second stage is the validation. In this stage, we evaluate each QA pair using two checks. First \goldcircled{5} we leverage an LLM-as-a-judge to \textit{assess question quality}, including whether the answer is correct, whether the question is grounded in the provided context, and whether the question is answerable. We also verify that the question requires all provided supporting contexts rather than only a subset of them. 
The second check \goldcircled{6} applies the notion of \textit{sufficient context} \cite{joren_sufficient_2025}. It tests whether any single provided context is sufficient to answer the question. If a question can be answered using only one context document, it is not a true multi-hop question and is therefore rejected.
These checks enable a \textit{feedback loop} in which the LLM generates feedback whenever a requirement is not satisfied \goldcircled{7a}. The generator then uses this feedback to revise the question, which is subsequently re-evaluated by the validation step (Examples in Appendix \ref{app:validation}).
All validation checks are mandatory to ensure high-quality multi-hop questions.
\subsection{Extraction}
The final stage is extraction. This stage is required both to simulate unanswerable questions and to enable further experiments on generator performance, while also introducing controllable difficulty levels for the retriever component.

First, the extraction stage augments each question with additional non-relevant contexts. For each relevant context, we \textit{retrieve similar contexts} from the corpus using similarity search \goldcircled{7b}. After collecting these candidate documents, we \textit{assess the relevance} of each candidate context \goldcircled{8}. For this step, we adapt the CARE approach by \citet{brehme_evaluating_2026}. Specifically, we provide the LLM with the set of known relevant contexts together with a new candidate context and ask whether the candidate is relevant. This yields a context list with relevance labels.

To derive a \textit{difficulty level}, we count the number of contexts labeled as relevant \goldcircled{9}. A low number of relevant contexts corresponds to high difficulty, because fewer useful chunks are available for answering the question. A high number corresponds to low difficulty, with medium difficulty in between. This difficulty signal helps to analyze retriever performance on specific question types and identify optimization potential.

To simulate unanswerable questions, we modify the context list by keeping only non-relevant contexts. This modified list replaces the retriever output and is passed to the generator as retrieved documents, allowing us to test whether the model correctly detects unanswerable questions.

\section{Experimental Setup}

In this section, we describe the construction of the experimental QA datasets using TRIAD, the RAG setups, the human validation procedure, the comparison QA datasets, and the evaluation across seven different RAG configurations.
All data, prompts (also in Appendix \ref{app:promots}), and code are available in our GitHub repository \cite{githubGitHubLorenzbrehmetriad}.

\subsection{Dataset generation}
For dataset generation, we used the full set of context documents from the HotpotQA and MuSiQue datasets as the source corpus. We used Gemini-2.5-Flash \cite{noauthor_gemini_nodate} as the generation model, Gemini-3.1-Flash-lite-Preview \cite{noauthor_gemini-3-1-flash-lite-model-card_nodate} as the validation model, and Gemini-2.5-Flash-Lite \cite{noauthor_gemini_nodate-1} as the extraction model.
For the vector store, we used a PGVector database ~\cite{noauthor_langchain-ailangchain-postgres_2025} and embedded the chunks exactly as in the original database using the BAAI/bge-small-en-v1.5 embedding model \cite{noauthor_baaibge-small-en-v15_2026}. To retrieve the top-$k$ most similar documents, we used the similarity search function.
For our experiments, we generated 600 questions with our approach, resulting in 478 valid questions for the MuSiQue corpus and 481 valid questions for the HotpotQA corpus after validation. During generation, we sampled 200 questions for each of the three multi-hop question types: 100 binary questions and 100 non-binary questions. 

\subsection{Human Validation}
To validate our QA set, we perform a manual evaluation on 200 questions, sampled evenly (100 each) from the two selected QA datasets, across four metrics. The first metric was answerability, which measures whether the answer to the question is present in the provided documents. In addition, we evaluated correctness, which measures whether the answer, or the ground-truth answer, is correct. The third metric was true multi-hop reasoning, which measures whether the question can be answered using a single chunk or whether all supporting chunks are required. The final metric was unambiguity, which measures whether the question can be answered unambiguously and has exactly one correct answer.
Additionally, we sampled 20 instances from each QA set, for a total of 40, to evaluate the validator in cases where validation failed and feedback was provided for regeneration. We then assessed whether this feedback was correct.

\subsection{Comparison QA Datasets}
To validate the QA datasets generated with TRIAD, we compared RAG performance on our datasets against existing QA benchmarks. Specifically, we used the MuSiQue and HotpotQA datasets and sampled 600 questions from each.
For the HotpotQA sample, we selected 100 questions from each class, where the classes were defined by question type (bridging vs. comparison) and difficulty level (easy, medium, and hard), resulting in six classes with 100 questions each. MuSiQue does not provide an equivalent class structure, so we sampled 600 questions uniformly at random.

\subsection{RAG-Setups}
We conduct our performance comparison using seven distinct RAG setups. We varied three components of the pipeline. First, we evaluated two different retrieval settings with 3 and 5 retrieved chunks. Second, we compared three embedding models: Google Embedding-001, Google Embedding-002 \cite{noauthor_embeddings_nodate}, and sentence-transformers/all-MiniLM-L6-v2 \cite{noauthor_sentence-transformersall-minilm-l6-v2_2024}. Third, we evaluated two LLMs, GPT-5-nano \cite{openAI_gpt-5_2025} and Gemma-4-31B-It \cite{noauthor_modellkarte_nodate}. For all setups, we use a PGVector database~\cite{noauthor_langchain-ailangchain-postgres_2025}.

We compared all of these configurations against a standard setup consisting of the BAAI/bge-small-en-v1.5 embedding model, 5 retrieved chunks, and Gemini-2.5-Flash-Lite. In each experiment, we varied one component at a time and analyzed whether our generated datasets exhibited performance trends similar to those observed on established benchmarks.

\subsection{RAG-Evaluation}
To evaluate the different RAG setups, we used four metrics. The first two were standard RAGAS metrics: faithfulness and answer correctness \cite{es_ragas_2023}. We evaluated both metrics using GPT-5-nano \cite{openAI_gpt-5_2025} in the default configuration.
The other two metrics were context precision and context recall. For these, we used the relevance labels provided by the datasets and checked whether the RAG system retrieved the same contexts using exact matching. We then computed precision and recall based on these matches.

\section{Validation}
This section presents the results of our validation experiments. We generate two QA datasets and first validate them through human annotation. We then assess their utility by comparing them with established multi-hop QA benchmarks and examining whether they yield similar performance trends across seven distinct RAG setups. Finally, we analyze the performance of different question types across these setups to demonstrate the effectiveness of the proposed QA categories. We additionally evaluate unanswerable questions to test the ability of RAG systems to detect when a query cannot be answered.

\subsection{Human Validation}

\begin{table}[ht]
\centering
\small
\begin{tabular}{p{1.0cm}p{1.2cm}p{1.2cm}p{1.2cm}p{1.2cm}}
\toprule
\textbf{Dataset} & \textbf{Answer-ability} & \textbf{Correct-ness} & \textbf{Multi-hop} & \textbf{Un-ambiguity} \\
\midrule
HotpotQA & 0.90 & 0.92 & 0.91 & 0.92 \\
MuSiQue & 0.93 & 0.98 & 0.85 & 0.92 \\
\bottomrule
\end{tabular}
\caption{Human validation results for TRIAD QA sets.}
\label{tab:human_validation_generated}
\end{table}

To assess the quality of the generated questions, we first examined a sample of questions accepted by the validator. We assessed their overall quality and checked whether the validator incorrectly accepted low-quality questions. We then examined a sample of rejected questions to determine how many valid, high-quality questions were incorrectly rejected.

For the accepted questions, Table \ref{tab:human_validation_generated} shows that 90\% of the questions are answerable, more than 92\% of the generated QA pairs are correct, and over 85\% require multi-hop reasoning. In addition, more than 92\% of the questions are unambiguous, indicating that the generated QA pairs are suitable for evaluating RAG systems. 

To further validate the behavior of the validator, we first measured how many of the generated QA instances were ultimately accepted. The feedback loop allowed for two iterations, meaning that a question was discarded if it was rejected twice. Each iteration was counted as one generated question. For HotpotQA, we generated a total of 703 questions. In total, 135 questions were rejected by the quality validator and 97 by the sufficient-context validator, resulting in 481 accepted questions and an acceptance rate of 68\%. For the MuSiQue dataset, we generated 723 questions under the same setup. Of these, 186 were rejected by the quality validator and 74 by the sufficient-context check, resulting in 478 accepted questions and an acceptance rate of 66\%.
We sampled 40 validation decisions, with 20 from each dataset, and manually checked whether the rejection reasons were correct. This yielded a true rejection rate of 90\% for MuSiQue and 85\% for HotpotQA. Thus, less than 15\% of the rejected instances were actually valid questions. 

\subsection{RAG-Evaluation}
To verify that our generation approach can be used to evaluate RAG systems, we evaluated seven distinct RAG setups and analyzed their performance on our generated datasets in comparison with the established HotpotQA and MuSiQue datasets.

\begin{figure}[ht]
    \centering

    \begin{subfigure}{0.48\textwidth}
        \centering
        \includegraphics[width=\linewidth]{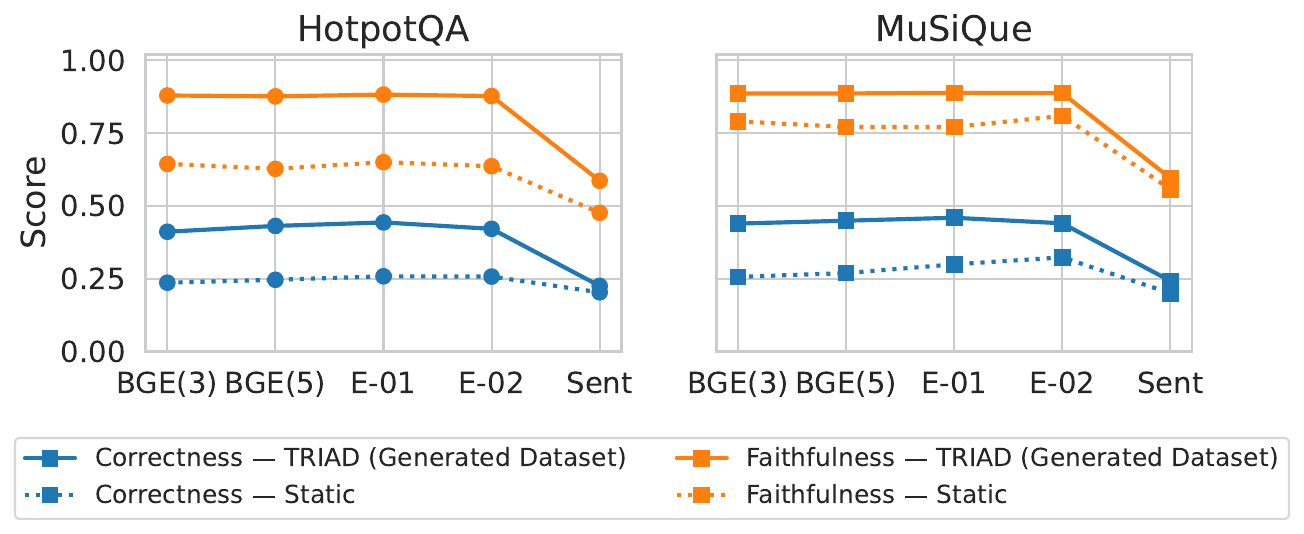}
        \caption{Generator metrics}
        \label{fig:embedding-generator-metrics}
    \end{subfigure}
    \hfill
    \begin{subfigure}{0.48\textwidth}
        \centering
        \includegraphics[width=\linewidth]{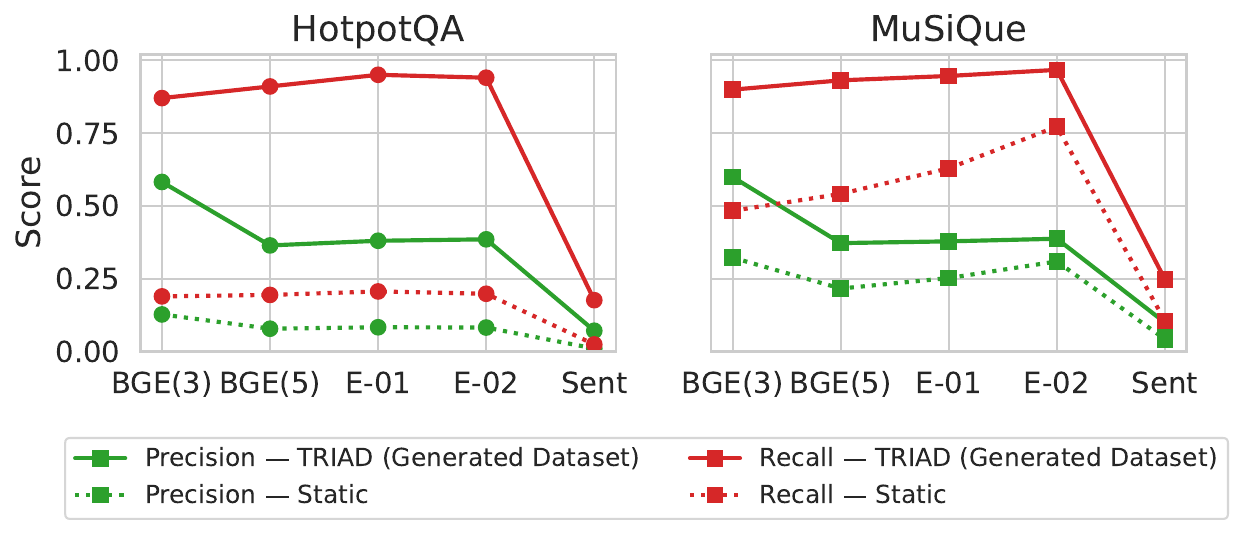}
        \caption{Retriever metrics}
        \label{fig:embedding-retriever-metrics}
    \end{subfigure}

    \caption{Embedding-model performance using Gemini-2.5-Flash-Lite (Tabular results Appendix \ref{app:results}).}
    \label{fig:embedding-comparison}
\end{figure}

We first varied the embedding model (cf. Figure \ref{fig:embedding-comparison}). Across all settings, the retriever performed best with Google's Embedding-002 model, followed by Embedding-001 \cite{noauthor_embeddings_nodate}, while sentence-transformers/all-MiniLM-L6-v2 \cite{noauthor_sentence-transformersall-minilm-l6-v2_2024} performed worst in terms of context recall. This trend was also reflected in context precision, answer correctness, and faithfulness. The strongest effects were observed for the retriever-focused metrics. In both datasets, the generated questions yielded substantially better retrieval performance. For HotpotQA, the recall difference exceeded 70\%, whereas for MuSiQue, the gap was smaller, remaining below 45\%.
Next, we reduced the number of retrieved chunks from 5 to 3. This change increased precision but decreased recall. Again, all datasets showed similar trends. The effect was least pronounced for the existing HotpotQA dataset, whereas the generated datasets exhibited larger improvements in precision and larger drops in recall. For example, context precision on HotpotQA increased from 0.078 to 0.127, whereas on the generated dataset it increased from 0.364 to 0.582.

\begin{table}[ht]
\centering
\small

\begin{tabular}{p{2cm}p{1.3cm}p{1.3cm}p{1.3cm}}
\toprule
\textbf{Metric} & \textbf{Gemini} & \textbf{Gemma} & \textbf{GPT-5} \\
\midrule

\multicolumn{4}{c}{\textbf{HotpotQA -- TRIAD (Generated Dataset)}} \\
\midrule
Correctness  & \textbf{0.431} & 0.387$\downarrow$ & 0.447$\uparrow$ \\
Faithfulness & \textbf{0.876} & 0.874 & 0.886$\uparrow$ \\

\midrule

\multicolumn{4}{c}{\textbf{HotpotQA -- Static}} \\
\midrule
Correctness  & \textbf{0.246} & 0.231$\downarrow$ & 0.309$\uparrow$ \\
Faithfulness & \textbf{0.627} & 0.563$\downarrow$ & 0.722$\uparrow$ \\

\midrule

\multicolumn{4}{c}{\textbf{MuSiQue -- TRIAD (Generated Dataset)}} \\
\midrule
Correctness  & \textbf{0.449} & 0.406$\downarrow$ & 0.459$\uparrow$ \\
Faithfulness & \textbf{0.886} & 0.886 & 0.883 \\

\midrule

\multicolumn{4}{c}{\textbf{MuSiQue -- Static}} \\
\midrule
Correctness  & \textbf{0.269} & 0.259$\downarrow$ & 0.374$\uparrow$ \\
Faithfulness & \textbf{0.771} & 0.571$\downarrow$ & 0.700$\downarrow$ \\

\bottomrule

\end{tabular}
\caption{Comparison of LLMs using the BGE(5) retriever. Arrows indicate improvements relative to the baseline. The baseline is shown in bold.}
\label{tab:llm_comparison}
\end{table}

As a final modification, we varied the LLM (cf. Table \ref{tab:llm_comparison}). This affected only faithfulness and answer correctness, while context precision and context recall remained unchanged. For answer correctness, GPT-5-nano achieved the best performance, and the overall trends were similar, with Gemma-4-31B-It performing worst.
The results for faithfulness were less consistent. On the generated dataset, all models showed relatively similar performance, whereas the existing datasets exhibited larger differences in faithfulness. The outcomes were not fully consistent across benchmarks. For example, Gemini-2.5-Flash-Lite achieved the highest faithfulness on MuSiQue, whereas GPT-5-nano performed best on HotpotQA.

\subsection{Question Types}
\begin{table}[ht]
\centering
\small
\begin{tabular}{p{1.7cm}p{1cm}p{1cm}p{0.9cm}p{0.9cm}}
\toprule
\textbf{Type} & \textbf{Correct} & \textbf{Faithful} & \textbf{Precision} & \textbf{Recall} \\
\midrule
Comparison* & 0.289 & 0.870 & 0.391 & 0.882 \\
Comparison & 0.390 & 0.895 & 0.387 & 0.873 \\
Intersection* & 0.383 & 0.817 & 0.345 & 0.782 \\
Intersection & 0.583 & 0.807 & 0.357 & 0.809 \\
Composition* & 0.318 & 0.832 & 0.356 & 0.806 \\
Composition & 0.478 & 0.823 & 0.343 & 0.778 \\
\bottomrule
\end{tabular}
\caption{Performance by question type.  * indicates binary questions.}
\label{tab:performance_per_type}
\end{table}

We also investigated performance across the different question types and calculated the average metric values across all datasets. Intersection questions achieved the highest answer correctness, whereas comparison questions obtained the lowest correctness scores. For faithfulness, comparison questions performed best, while intersection and composition questions scored lower. A similar pattern was observed for context precision and context recall. For binary and non-binary questions, the non-binary variants achieved substantially higher answer correctness than the binary ones. For the other metrics, a slight trend was observable. The differences were small, with a maximum gap of 2.5\%. In six cases, binary questions performed better, whereas in three cases, non-binary questions performed better.

\subsection{Unanswerable Questions}
\label{sec:unanswerable}
To evaluate unanswerable questions, we used the three models from the previous experiments and provided each of them with five non-relevant context documents. We then measured the rate at which the models correctly detected that a question could not be answered from the given context. Specifically, we checked whether the model stated that the provided context did not contain sufficient information. Gemini-2.5-Flash-Lite and Gemma-4-31B-It identified such cases correctly in more than 99\% of the questions, whereas GPT-5-nano achieved a detection rate of 94\%. In cases where the model failed to detect that the provided context was insufficient, it hallucinated an answer based on prior knowledge, even when that answer happened to be correct. We counted such cases as incorrect detections.

\section{Discussion}
Our results demonstrate that TRIAD can be used to create multi-hop questions from an existing vector database that is already used by the RAG system. This may help address a major obstacle to the practical adoption of RAG evaluation in industry, namely the complexity and time required to create high-quality evaluation datasets \cite{brehme_retrieval-augmented_2026}. Both the human validation and the experimental results indicate that the proposed approach is a viable method for evaluating and comparing different RAG setups using a generated QA set.

We also address the issue of unanswerable questions and provide a method for benchmarking how often RAG systems hallucinate answers. In our experiments, GPT-5-nano exhibited the highest hallucination rate among the tested models. To measure this behavior, we simulated retrieval failure by providing only non-relevant context documents for a question. In such cases, the model should explicitly state that the answer cannot be derived from the provided context.

In addition, we introduced different question types that are designed to reflect different retrieval strategies. This enables a more fine-grained analysis than relying only on a single aggregate score. As a result, the benchmark can reveal which question types a RAG system handles particularly well. If the expected user needs are known in advance, practitioners can place greater emphasis on the most relevant question types and construct a benchmark that is better aligned with the intended application.

Our analysis of the validator revealed a false rejection rate between 10\% and 15\%, depending on the dataset. We intentionally preferred stricter rejection criteria, accepting that some good questions would be discarded rather than retaining too many low-quality ones. For this reason, we designed the validator to apply several checks and a feedback loop to ensure that the final QA set is of high quality and suitable for evaluation.

A potential concern is that the generated questions may be easier to retrieve than those in human-created datasets, as reflected in the higher absolute scores. This effect was particularly evident in the retriever metrics, especially context precision and context recall. One likely explanation lies in the question generation process itself: because it relies on identifying semantically similar documents, retrievers based on semantic similarity may find it easier to retrieve the relevant contexts. At the same time, the relative performance trends across the evaluated RAG setups remained consistent. This suggests that the generated datasets can still support meaningful comparative evaluation, even if their retrieval difficulty differs from that of established benchmarks.

Another concern is that the performance of the evaluated RAG systems may be biased when both question generation and downstream evaluation rely on the same vector database and similar retriever models. Since the questions are generated from retrieved documents, systems using comparable retrieval strategies could, in principle, gain an advantage. To investigate this possibility, we compared different embedding models and analyzed the resulting performance trends. We observed patterns similar to those found on established benchmarks, suggesting that the results are not primarily driven by bias introduced during the generation phase.

\section{Conclusion}
In summary, this paper introduces TRIAD, a new approach for generating a domain-specific multi-hop dataset for evaluating RAG systems. The approach defines question types that are designed to trigger different retrieval patterns. The resulting dataset contains question--answer pairs, relevant contexts for answering each question, and distractor contexts that can be used to simulate unanswerable questions and to modify retriever outputs during evaluation. The validation results show that the approach is suitable for evaluating RAG systems. Although the generated dataset is still weaker than human-created benchmarks such as HotpotQA, it shows similar performance trends across different RAG setups. This suggests that the approach can be used for comparative RAG evaluation in domain-specific settings.
Future work includes extending the framework to multi-turn, multi-hop question answering and evaluating the dataset in real-world settings with domain expert validation to assess the realism and practical relevance of the generated questions.

\section{Limitations}

This section discusses the limitations of this paper. First, the reported results are limited to the models and prompts used in our experiments and may not generalize to other models or prompt settings. In addition, because the study relies on LLMs, the experiments are not fully reproducible due to their non-deterministic behavior. We provide all data and code in our GitHub repository and describe the experimental setup in as much detail as possible. However, exact replication of the results may still not be possible, even when using the same settings on the same machine.
Additionally, the correctness and faithfulness metrics rely on an LLM as a judge and therefore on the reliability of both the model and the RAGAS framework \cite{es_ragas_2023}, an established evaluation framework used in prior work. The other metrics were either based on exact matches or assessed by humans.

More generally, the approach was evaluated on relatively clean corpora and mostly general-knowledge data. Future work should therefore test the method in an industry use case, including noisier data and more domain-specific knowledge on which the model has not been explicitly trained. This would allow for a broader assessment of the approach's performance.

\section*{Acknowledgements}
The authors used ChatGPT, Grammarly, and GitHub Copilot for language, grammar, and coding support. All content was reviewed and verified by the authors.
\bibliography{custom}

 \appendix
\section*{Appendix}
The following appendices provide supplementary material for the paper. Section~\ref{app:results} presents a table corresponding to the results shown in Figure~\ref{fig:embedding-comparison}, followed by detailed results for Section~\ref{sec:unanswerable}. Section~\ref{app:validation} provides three examples illustrating the feedback loop, and Section~\ref{app:questions} contains questions generated using the TRIAD approach. Finally, Section~\ref{app:promots} includes all prompts used throughout the experiments.
\section{Results}
\label{app:results}
\begin{table}[H]
\centering
\small

\begin{tabular}{p{1.2cm}p{0.8cm}p{0.7cm}p{0.8cm}p{0.8cm}p{0.8cm}}
\toprule \textbf{Metric} & \textbf{BGE(3)} & \textbf{BGE(5)} & \textbf{Sent} & \textbf{E-01} & \textbf{E-02} \\
\midrule

\multicolumn{6}{c}{\textbf{HotpotQA -- Generated}} \\
\midrule
Correctness  & 0.411$\downarrow$ & \textbf{0.431} & 0.225$\downarrow$ & 0.443$\uparrow$ & 0.421$\downarrow$ \\
Faithfulness & 0.879 & \textbf{0.876} & 0.586$\downarrow$ & 0.882 & 0.877 \\
Precision    & 0.582$\uparrow$ & \textbf{0.364} & 0.071$\downarrow$ & 0.380$\uparrow$ & 0.385$\uparrow$ \\
Recall       & 0.870$\downarrow$ & \textbf{0.910} & 0.176$\downarrow$ & 0.950$\uparrow$ & 0.940$\uparrow$ \\
\midrule

\multicolumn{6}{c}{\textbf{HotpotQA -- Static}} \\
\midrule
Correctness  & 0.236$\downarrow$ & \textbf{0.246} & 0.203$\downarrow$ & 0.258$\uparrow$ & 0.257$\uparrow$ \\
Faithfulness & 0.644$\uparrow$ & \textbf{0.627} & 0.476$\downarrow$ & 0.650$\uparrow$ & 0.636\\
Precision    & 0.127$\uparrow$ & \textbf{0.078} & 0.010$\downarrow$ & 0.083$\uparrow$ & 0.082$\uparrow$ \\
Recall       & 0.189$\downarrow$ & \textbf{0.194} & 0.024$\downarrow$ & 0.206 & 0.198$\uparrow$ \\
\midrule

\multicolumn{6}{c}{\textbf{MuSiQue -- Generated}} \\
\midrule
Correctness  & 0.439$\downarrow$ & \textbf{0.449} & 0.242$\downarrow$ & 0.459$\uparrow$ & 0.440 \\
Faithfulness & 0.886 & \textbf{0.886} & 0.595$\downarrow$ & 0.888 & 0.887 \\
Precision    & 0.599$\uparrow$ & \textbf{0.372} & 0.099$\downarrow$ & 0.378 & 0.387$\uparrow$ \\
Recall       & 0.899$\downarrow$ & \textbf{0.931} & 0.247$\downarrow$ & 0.946$\uparrow$ & 0.967$\uparrow$ \\
\midrule

\multicolumn{6}{c}{\textbf{MuSiQue -- Static}} \\
\midrule
Correctness  & 0.256$\downarrow$ & \textbf{0.269} & 0.200$\downarrow$ & 0.299$\uparrow$ & 0.323$\uparrow$ \\
Faithfulness & 0.790$\uparrow$ & \textbf{0.771} & 0.557$\downarrow$ & 0.771 & 0.809$\uparrow$ \\
Precision    & 0.322$\uparrow$ & \textbf{0.216} & 0.042$\downarrow$ & 0.252$\uparrow$ & 0.309$\uparrow$ \\
Recall       & 0.483$\downarrow$ & \textbf{0.541} & 0.105$\downarrow$ & 0.629$\uparrow$ & 0.772$\uparrow$ \\
\bottomrule

\end{tabular}
\caption{Comparison of embedding models using Gemini-2.5-Flash-Lite (cf. Figure \ref{fig:embedding-comparison}). Arrows indicate improvements relative to the baseline. The baseline is shown in bold.}
\label{tab:embedding_comparison}
\end{table}

\begin{table}[H]
    \centering
    \begin{tabular}{p{1.5cm}p{1.5cm}p{1.5cm}p{1.5cm}}
    \toprule
       \textbf{Dataset} & \textbf{Gemini} & \textbf{Gemma} & \textbf{GPT-5} \\
       \midrule
       MuSiQue  & 99.59\% & 99.58\% & 93.78\%\\
       \midrule
       HotpotQA  & 99.58\% & 99.37\% & 94.76\%\\
       \bottomrule
    \end{tabular}
    \caption{Comparison of scores for unanswerable questions (cf. Section~\ref{sec:unanswerable}). }
    \label{tab:my_label}
\end{table}

\section{Example of Validation Feedback}
\label{app:validation}

\begin{tcolorbox}[
    colback=gray!5,
    colframe=gray!75,
    title=Non-Standalone Question,
    breakable,
    enhanced
]
\small
\centering

\begin{minipage}{0.95\linewidth}
\textbf{Rejected Question:} \\
Which animal breed, the Romanian Raven Shepherd Dog or the Tsigai, is explicitly stated in the provided text as originating from Asia Minor?
\end{minipage}

\vspace{0.5em}

\begin{minipage}{0.95\linewidth}
\textbf{Rejection Reason:} \\
The question is of low quality for a RAG system because it is not standalone and explicitly references “the provided text”. It is also ambiguous, since only the Tsigai origin is stated while the Romanian Raven Shepherd Dog origin is not provided, making the comparison unsupported. A clearer reformulation would be: “Where did the Tsigai sheep breed originate?”
\end{minipage}
\vspace{0.5em}

\begin{minipage}{0.95\linewidth}
\textbf{Improved Question:} \\
Which of these breeds, the Romanian Raven Shepherd Dog or the Tsigai, is explicitly identified as the leading breed in Moldova?
\end{minipage}
\label{fig:rejected-question-breed-origin}
\end{tcolorbox}

\begin{tcolorbox}[
    colback=gray!5,
    colframe=gray!75,
    title=Ambiguous Question,
    breakable,
    enhanced
]
\small
\centering

\begin{minipage}{0.95\linewidth}
\textbf{Rejected Question:} \\
Did the actress who received an Academy Award nomination for Best Supporting Actress also win four Academy Awards for Best Actress?
\end{minipage}

\vspace{0.5em}

\begin{minipage}{0.95\linewidth}
\textbf{Rejection Reason:} \\
The question is of low quality for a RAG system because it fails the 'Standalone Retrieval Clarity' and 'Explicit Entity Grounding' criteria. The question uses the phrase 'the actress who received an Academy Award nomination for Best Supporting Actress' without naming her. Because there could be many actresses who have received such a nomination, the retriever would not know which specific actress or documents to target, forcing the system to rely on the context to identify the subject rather than the question itself. To improve this, the question should explicitly name the actress: 'Did Katharine Ross, who received an Academy Award nomination for Best Supporting Actress, also win four Academy Awards for Best Actress?'
\end{minipage}
\vspace{0.5em}

\begin{minipage}{0.95\linewidth}
\textbf{Improved Question:} \\
Was the actress noted for her role as Elaine Robinson in "The Graduate" also recognized by the American Film Institute as the greatest female star of Classic Hollywood Cinema?
\end{minipage}
\label{fig:rejected-question-entity-grounding}
\end{tcolorbox}

\begin{tcolorbox}[
    colback=gray!5,
    colframe=gray!75,
    title=Unanswerable Question,
    breakable,
    enhanced
]
\small
\centering

\begin{minipage}{0.95\linewidth}
\textbf{Rejected Question:} \\
Did the founding of Epicureanism by Epicurus occur earlier than Democritus's coining of atomic theory?
\end{minipage}

\vspace{0.5em}

\begin{minipage}{0.95\linewidth}
\textbf{Rejection Reason:} \\
The question is of low quality for RAG because it is not fully answerable based solely on the provided context. While the context provides the founding date of Epicureanism (307 BC), it mentions that Democritus coined atomic theory but provides no timeframe or date for this event. Therefore, it is impossible to compare the two chronologically using only the provided text. Additionally, because the answer cannot be derived, the question is ambiguous and fails the requirement for factual grounding.
\end{minipage}
\vspace{0.5em}

\begin{minipage}{0.95\linewidth}
\textbf{Improved Question:} \\
Did Epicurus's atomic materialism, which followed Democritus, consider the individual as the 'atom' of society in the same way sociological atomism does?
\end{minipage}
\end{tcolorbox}

\section{Examples of Questions}
\label{app:questions}

\begin{table*}[ht]
\small
\centering
\begin{tabular}{p{7cm} p{2cm} p{1.5cm} p{1cm}p{1.5cm}}
\midrule
\textbf{Question} & \textbf{Answer} & \textbf{Type} & \textbf{Binary} & \textbf{Corpus}\\
\midrule
Did Hijokaidan's early lineup have more members than the Doberman ensemble? & Yes & Comparison & Yes & HotpotQA\\
\midrule
Did the British Interplanetary Society's study for Project Daedalus last longer than their study for Project Boreas? & Yes & Comparison  & Yes & MuSiQue \\
\midrule
Which single, Will Young's 'Love Revolution' or Tomcraft's 'Loneliness', was released earlier in the United Kingdom? & Tomcraft's 'Loneliness & Comparison  & No & HotpotQA \\
\midrule
Who was born earlier, Alejandro Llorente or Don Carlos, Duke of Madrid? & Alejandro Llorente & Comparison  & No & MuSiQue \\
\midrule
Did both the 17/18 and the 16/17 UEFA Champions League knockout phases include a total of 16 teams? & Yes & Intersection & Yes & HotpotQA\\
\midrule
Is there an actress mentioned who played both a love interest of Ted Mosby and the first lead female character for the Call of Duty franchise? & No & Intersection  & Yes & MuSiQue \\
\midrule
Which individual both produced The Matches' album Decomposer and owns Epitaph Records? & Brett Gurewitz & Intersection & No & HotpotQA\\
\midrule
Which constellation contains a white dwarf star that shows traces of external metal contamination and also a G-type giant star approximately 331 light years away? & Cetus & Intersection  & No & MuSiQue \\
\midrule
Was the fourth studio album by the band for which Peter Dennis Blandford Townshend is the main songwriter first released in 1969? & Yes & Composition & Yes & HotpotQA\\
\midrule
Was the philosopher whose collected essays were published as 'The Windmills of Humanity' associated with the Frankfurt School? & No & Composition  & Yes & MuSiQue \\
\midrule
What company produced the science fiction film that Bonnie MacBird was the original writer of? & Walt Disney Productions & Composition & No & HotpotQA\\
\midrule
Who rules as king of the gods to whom ambrosia was brought by doves? & Zeus & Composition  & No & MuSiQue \\
\midrule
\end{tabular}
\caption{Example multi-hop QA samples from our dataset.}
\label{tab:qa_samples}
\end{table*}

\newpage
\section{Prompts}
\newpage
\label{app:promots}
\subsection{Generation Stage}
\lstset{
    basicstyle=\ttfamily\small,
    breaklines=true,
    breakatwhitespace=true,
    columns=fullflexible
}

\begin{tcolorbox}[
    colback=gray!5,
    colframe=gray!75,
    title=Question Generation Prompt (Comparison Question),
    breakable,
    enhanced
]
\small
\begin{lstlisting}
You are an expert at generating high-quality multi-hop questions for Retrieval-Augmented Generation (RAG) evaluation.

You will be given multiple text chunks containing factual information.

Your task is to generate:
1. A multi-hop question
2. Its short factual answer

The generated question will be used as a USER QUERY in a RAG system.

IMPORTANT RAG CONSTRAINTS
- The retriever only sees the question.
- The retriever does NOT initially know the chunks.
- Therefore, the question must contain enough explicit information to retrieve the relevant chunks.
- The question must be understandable WITHOUT access to the chunks.
- Avoid vague references or hidden bridge entities unless explicitly required by the reasoning type.

-------------------------------------
GLOBAL REQUIREMENTS
-------------------------------------

The question MUST:

1. Require multi-hop reasoning
- The answer must require combining information from AT LEAST TWO chunks.
- No single chunk may fully answer the question.

2. Be concise and clear
- Prefer a single sentence.
- Avoid unnecessary wording.

3. Include retrieval anchors
- Include at least one explicit named entity
  (person, company, publication, event, location, etc.)
  unless the reasoning type explicitly forbids it.

4. Avoid answer leakage
- Do NOT reveal the answer in the question.
- Do NOT copy phrases that trivially expose the answer.

5. Produce a short factual answer
- The answer should be:
  - a name
  - a number
  - a date
  - a location
  - a short phrase
  - or a short list

6. Be uniquely answerable
- The question must have EXACTLY ONE correct answer.
- Avoid ambiguity or multiple valid interpretations.

7. Be fully grounded in the chunks
- Do NOT use external knowledge.
- Do NOT invent facts.
- Do NOT infer unstated relationships.

8. Pass the necessity test
- Chunk A alone must NOT answer the question.
- Chunk B alone must NOT answer the question.
- Combining chunks MUST be necessary.

-------------------------------------
REASONING TYPE
-------------------------------------

Reasoning Type:
Comparison

Reasoning Instructions:
Generate a comparison-based multi-hop question. The question should: - retrieve two entities or fact sets from different chunks, - compare them along a specific attribute, - and determine the final answer from the comparison. The comparison may involve: - dates - sizes - rankings - locations - quantities - roles - durations - achievements - or other factual attributes. Requirements: - Both compared entities must come from different chunks. - The comparison result must not be directly stated anywhere. - The answer must require combining and comparing information. Good example: "Which scientist received their Nobel Prize earlier, Marie Curie or Niels Bohr?"

-------------------------------------
INPUT
-------------------------------------

Text Chunks:
{chunks}

Bridge Topic:
{bridge_topic}

Role:
{role}

-------------------------------------
OUTPUT FORMAT
-------------------------------------

Respond with VALID JSON ONLY:

{{
  "multi_hop_question": "<generated question>",
  "multi_hop_answer": "<short factual answer>"
}}
{binary}

\end{lstlisting}
\end{tcolorbox}

\begin{tcolorbox}[
    colback=gray!5,
    colframe=gray!75,
    title=Question Generation Prompt (Composition Question),
    breakable,
    enhanced
]
\small
\begin{lstlisting}
You are an expert at generating high-quality multi-hop questions for Retrieval-Augmented Generation (RAG) evaluation.

You will be given multiple text chunks containing factual information.

Your task is to generate:
1. A multi-hop question
2. Its short factual answer

The generated question will be used as a USER QUERY in a RAG system.

IMPORTANT RAG CONSTRAINTS
- The retriever only sees the question.
- The retriever does NOT initially know the chunks.
- Therefore, the question must contain enough explicit information to retrieve the relevant chunks.
- The question must be understandable WITHOUT access to the chunks.
- Avoid vague references or hidden bridge entities unless explicitly required by the reasoning type.

-------------------------------------
GLOBAL REQUIREMENTS
-------------------------------------

The question MUST:

1. Require multi-hop reasoning
- The answer must require combining information from AT LEAST TWO chunks.
- No single chunk may fully answer the question.

2. Be concise and clear
- Prefer a single sentence.
- Avoid unnecessary wording.

3. Include retrieval anchors
- Include at least one explicit named entity
  (person, company, publication, event, location, etc.)
  unless the reasoning type explicitly forbids it.

4. Avoid answer leakage
- Do NOT reveal the answer in the question.
- Do NOT copy phrases that trivially expose the answer.

5. Produce a short factual answer
- The answer should be:
  - a name
  - a number
  - a date
  - a location
  - a short phrase
  - or a short list

6. Be uniquely answerable
- The question must have EXACTLY ONE correct answer.
- Avoid ambiguity or multiple valid interpretations.

7. Be fully grounded in the chunks
- Do NOT use external knowledge.
- Do NOT invent facts.
- Do NOT infer unstated relationships.

8. Pass the necessity test
- Chunk A alone must NOT answer the question.
- Chunk B alone must NOT answer the question.
- Combining chunks MUST be necessary.

-------------------------------------
REASONING TYPE
-------------------------------------

Reasoning Type:
Attribute Compostion

Reasoning Instructions:
Generate a bridge-style multi-hop question. The question must require a two-step lookup: Step 1: Use one chunk to identify a hidden bridge entity. Step 2: Use the bridge entity to retrieve the final answer from another chunk. Critical Constraint: - The bridge entity MUST NOT appear explicitly in the question. - The bridge entity should only be recoverable through reasoning. Requirements: - Chunk A identifies the bridge entity. - Chunk B contains the final answer related to that entity. - Neither chunk alone should answer the question. Good example: "What is the capital of the country where the CEO of Company Y was born?"
-------------------------------------
INPUT
-------------------------------------

Text Chunks:
{chunks}

Bridge Topic:
{bridge_topic}

Role:
{role}

-------------------------------------
OUTPUT FORMAT
-------------------------------------

Respond with VALID JSON ONLY:

{{
  "multi_hop_question": "<generated question>",
  "multi_hop_answer": "<short factual answer>"
}}
{binary}

\end{lstlisting}
\end{tcolorbox}

\begin{tcolorbox}[
    colback=gray!5,
    colframe=gray!75,
    title=Question Generation Prompt (Intersection Question),
    breakable,
    enhanced
]
\small
\begin{lstlisting}
You are an expert at generating high-quality multi-hop questions for Retrieval-Augmented Generation (RAG) evaluation.

You will be given multiple text chunks containing factual information.

Your task is to generate:
1. A multi-hop question
2. Its short factual answer

The generated question will be used as a USER QUERY in a RAG system.

IMPORTANT RAG CONSTRAINTS
- The retriever only sees the question.
- The retriever does NOT initially know the chunks.
- Therefore, the question must contain enough explicit information to retrieve the relevant chunks.
- The question must be understandable WITHOUT access to the chunks.
- Avoid vague references or hidden bridge entities unless explicitly required by the reasoning type.

-------------------------------------
GLOBAL REQUIREMENTS
-------------------------------------

The question MUST:

1. Require multi-hop reasoning
- The answer must require combining information from AT LEAST TWO chunks.
- No single chunk may fully answer the question.

2. Be concise and clear
- Prefer a single sentence.
- Avoid unnecessary wording.

3. Include retrieval anchors
- Include at least one explicit named entity
  (person, company, publication, event, location, etc.)
  unless the reasoning type explicitly forbids it.

4. Avoid answer leakage
- Do NOT reveal the answer in the question.
- Do NOT copy phrases that trivially expose the answer.

5. Produce a short factual answer
- The answer should be:
  - a name
  - a number
  - a date
  - a location
  - a short phrase
  - or a short list

6. Be uniquely answerable
- The question must have EXACTLY ONE correct answer.
- Avoid ambiguity or multiple valid interpretations.

7. Be fully grounded in the chunks
- Do NOT use external knowledge.
- Do NOT invent facts.
- Do NOT infer unstated relationships.

8. Pass the necessity test
- Chunk A alone must NOT answer the question.
- Chunk B alone must NOT answer the question.
- Combining chunks MUST be necessary.

-------------------------------------
REASONING TYPE
-------------------------------------

Reasoning Type:
Intersection

Reasoning Instructions:
Generate a parallel multi-hop question. The question must require intersecting information from multiple chunks. Requirements: - Each chunk should contain multiple candidate entities. - The answer must be the ONLY entity satisfying conditions from BOTH chunks. - The reasoning should be symmetric. - Avoid sequential phrasing like: - "also" - "then" - "after identifying" Preferred forms: - "Which artist both X and Y?" - "Which organization appears in both..." - "Which athlete satisfies both conditions?" Critical Constraint: - Neither chunk alone should uniquely determine the answer. - Only the intersection should produce the answer.
-------------------------------------
INPUT
-------------------------------------

Text Chunks:
{chunks}

Bridge Topic:
{bridge_topic}

Role:
{role}

-------------------------------------
OUTPUT FORMAT
-------------------------------------

Respond with VALID JSON ONLY:

{{
  "multi_hop_question": "<generated question>",
  "multi_hop_answer": "<short factual answer>"
}}
{binary}
\end{lstlisting}
\end{tcolorbox}

\begin{tcolorbox}[
    colback=gray!5,
    colframe=gray!75,
    title=Bridging Topic Prompt,
    breakable,
    enhanced
]
\small
\begin{lstlisting}
You are given two text chunks.

Your task is to determine whether they contain at least one shared entity.

A shared entity is:
- The exact same named person, organization, location, event, work (TV show, book, movie), or other proper noun
- Or a clearly coreferent entity (e.g., "E! network" and "E!" count as the same entity)

Chunks: {chunks}

Instructions:
1. Extract all named entities from Chunk A.
2. Extract all named entities from Chunk B.
3. Compare the two lists.
4. Identify any shared entities.
5. Return your answer in the following JSON format:


{{  
"shared_entity_exists": true/false,
"shared_entities": [list of shared entities],
"entities_chunk_a": [list], 
"entities_chunk_b": [list]
}}

\end{lstlisting}
\end{tcolorbox}

\begin{tcolorbox}[
    colback=gray!5,
    colframe=gray!75,
    title=Role Prompt,
    breakable,
    enhanced
]
\small
\begin{lstlisting}
You are a question generator for a RAG (Retrieval-Augmented Generation) system focused on general knowledge about the world. Your task is to create short, precise, and unambiguous questions.
\end{lstlisting}
\end{tcolorbox}

\begin{tcolorbox}[
    colback=gray!5,
    colframe=gray!75,
    title=Temporal Classification Prompt,
    breakable,
    enhanced
]
\small
\begin{lstlisting}
Classify the following question as temporal or non-temporal.

Question:
{question}

Answer:
{answer}
Output format:
Respond with exactly one word:
Temporal
or
Non-Temporal
\end{lstlisting}
\end{tcolorbox}

\subsection{Validation Stage}

\begin{tcolorbox}[
    colback=gray!5,
    colframe=gray!75,
    title=Validation Check 1 Prompt,
    breakable,
    enhanced
]
\small
\begin{lstlisting}
You are evaluating a question for Retrieval-Augmented Generation (RAG).

  The question will be used as a USER QUERY in a RAG pipeline.

  IMPORTANT CONTEXT:
  - The final downstream system DOES NOT initially know the context chunks.
  - The system must retrieve relevant chunks ONLY from the question itself.
  - Therefore, the question must be understandable and meaningful WITHOUT access to the chunks.
  - The question should contain enough explicit information to retrieve the correct documents.
  - Questions that rely on hidden context, vague references, or implicit entities are LOW QUALITY for RAG.

  You are given:
  - A question
  - An answer
  - Multiple context passages (chunks)

  Your task is to evaluate the question across THREE dimensions:

  1. Question Quality
  2. Multi-hop Reasoning Requirement
  3. Unambiguity

  You must evaluate strictly using ONLY the provided context.

  -------------------------------------
  EVALUATION DIMENSIONS
  -------------------------------------

  1. QUESTION QUALITY

  Determine whether the question is high quality for RAG retrieval and answering.

  A high-quality RAG question MUST satisfy ALL of the following:

  A. Correctness
  - The question must be factually meaningful.
  - It must not contain contradictions, false assumptions, or nonsensical premises.

  B. Answerability from Context
  - The question must be fully answerable using ONLY the provided context.
  - All required entities, facts, and relationships must be explicitly present in the context.
  - Do NOT rely on external knowledge or unstated inference.

  C. Standalone Retrieval Clarity
  - The question must be understandable WITHOUT seeing the context.
  - The question must independently contain enough identifying information to retrieve the correct chunks.
  - It must NOT rely on hidden context for interpretation.

  Examples of BAD RAG questions:
  - Based on the provided context, which invention is attributed to Nikola Tesla?
  - According to the player profiles in the context, which footballer plays as a goalkeeper?
  - Why was the discovery of penicillin considered a major medical breakthrough in the provided material?
  - What event occurred immediately after the signing of the peace treaty described in the context?

  These are bad because the retriever would not know:
  - who "he" is
  - which company
  - what "it" refers to
  - which event is referenced

  Examples of GOOD RAG questions:
  - "What did Nikola Tesla invent after moving to the United States?"
  - "When did Apple launch the iPhone 14?"
  - "Why was the Apollo 11 moon landing historically important?"

  D. Explicit Entity Grounding
  - All critical entities must be explicitly named in the question itself.
  - Avoid unresolved pronouns or generic references:
  "he", "she", "they", "it", "this", "that", "the company", etc.
  - The question should be semantically searchable.

  Strict Rules
  - Do NOT assume missing information.
  - Do NOT use external knowledge.
  - Do NOT infer unstated relationships.
  - If the question depends on context to make sense, mark it invalid.
  - If the retriever could not reasonably retrieve the correct chunks using only the question, mark it invalid.
  - If the context is insufficient to fully answer the question, mark it invalid.
  - When uncertain, mark invalid.

  -------------------------------------

  2. MULTI-HOP REASONING

  Determine whether answering the question requires combining information from MULTIPLE chunks.

  Definition:
  A question is multi-hop if the answer requires combining information from TWO OR MORE different chunks.

  Procedure
  1. Identify all information required to answer the question.
  2. Identify which chunk(s) contain each required piece.
  3. Determine:
  - If ANY SINGLE chunk fully answers the question -> NOT multi-hop
  - If multiple chunks must be combined -> multi-hop

  Strict Rules
  - Do NOT use external knowledge.
  - Do NOT assume missing information.
  - If any single chunk alone is sufficient, mark as NOT multi-hop.
  - When uncertain, mark as NOT multi-hop.

  -------------------------------------

  3. UNAMBIGUITY

  Determine whether the question is unambiguous.

  Definition:
  A question is unambiguous if it has exactly ONE clear answer based ONLY on the provided context.

  Evaluation Criteria
  - The question must refer to clearly identifiable entities.
  - The question must not allow multiple interpretations.
  - The context must not support multiple valid answers.
  - All references must be explicitly clear within the question itself.

  Strict Rules
  - Do NOT use external knowledge.
  - Do NOT assume missing information.
  - Do NOT resolve ambiguity using context if the question itself is unclear.
  - If ANY ambiguity exists, mark ambiguous.
  - When uncertain, mark ambiguous.

  -------------------------------------
  OUTPUT FORMAT
  -------------------------------------

  Return ONLY valid JSON in the following format:

  {{
      "reason": "Detailed explanation of the final decision, referencing specific evaluation failures if rejected. And a suggestion for how to improve the question if it was rejected.",
      "accepted": true | false,
  }}

  -------------------------------------
  INPUT
  -------------------------------------

  Question:
  {question}

  Answer:
  {answer}

  Context:
  {contexts}

\end{lstlisting}
\end{tcolorbox}

\begin{tcolorbox}[
    colback=gray!5,
    colframe=gray!75,
    title=Validation Check 2 Prompt,
    breakable,
    enhanced
]
\small
\begin{lstlisting}
You are given a question and a context passage.

Your task is to determine whether the context ALONE contains ALL information required to answer the question completely and unambiguously.

Procedure:
1. Identify every piece of information required to answer the question (entities, relationships, attributes, dates, etc.).
2. Verify that EACH required element is explicitly stated in the context.
3. Verify that all relationships mentioned in the question (e.g., "sequel to", "author of", "capital of", "directed by") are explicitly stated in the context.
4. Check that the answer can be derived using only the context without any external knowledge.

Strict rules:
- Do NOT assume missing relationships.
- Do NOT rely on world knowledge.
- Do NOT infer unstated connections between entities.
- If the question references a relationship, the context must explicitly state that relationship.
- If even one required fact, entity link, or relationship is missing, respond "no".
- If the answer would require guessing or prior knowledge, respond "no".
- When uncertain, respond "no".

Output format:
Respond with exactly one word:
yes
or
no

Question: {question}
Context: {context}
\end{lstlisting}
\end{tcolorbox}

\subsection{Extraction Stage}

\begin{tcolorbox}[
    colback=gray!5,
    colframe=gray!75,
    title=Context Extraction Prompt,
    breakable,
    enhanced
]
\small
\begin{lstlisting}
Question : {question}
Answer : {answer}
Contextlist : {contexts}
Evaluate this context : {sentence}
Task : Does this sentence help support the answer , considering
the overall contextlist ?
Respond with :
" Relevant " or " Not Relevant " - nothing else .
\end{lstlisting}
\end{tcolorbox}

\end{document}